\documentclass[11pt]{article}

\usepackage[final]{arxiv}

\usepackage{times}
\usepackage{latexsym}

\usepackage[T1]{fontenc}

\usepackage[utf8]{inputenc}

\usepackage{microtype}

\usepackage{inconsolata}

\usepackage{url}
\usepackage{graphicx}
\usepackage{float}
\usepackage{placeins}
\usepackage{afterpage}
\usepackage{wrapfig}

\usepackage{array}
\usepackage{tabularx}
\usepackage{booktabs}
\usepackage[table]{xcolor}
\usepackage{multirow}
\usepackage{adjustbox}
\usepackage{makecell}
\usepackage{graphicx}

\usepackage{amsmath}
\usepackage{amssymb}
\usepackage{bm}

\usepackage{booktabs}      
\usepackage[table,svgnames]{xcolor}   
\usepackage{multirow}      
\usepackage{subcaption}
\usepackage{adjustbox}
\usepackage{tabularx}
\usepackage{makecell}

\usepackage{pifont}
\usepackage[ruled,vlined]{algorithm2e}
\usepackage[most]{tcolorbox}
\usepackage[inline]{enumitem}
\usepackage[T1]{fontenc}
\usepackage{fontawesome}
\usepackage{titletoc}

\definecolor{down}{HTML}{B1281C}
\definecolor{up}{HTML}{489638}
\definecolor{darkblue}{HTML}{1F33B4}

\newtcolorbox{templatebox}[1]{
    breakable,
    enhanced,
    colback=white,
    colframe=darkblue!80,
    colbacktitle=darkblue!80,
    coltitle=white,
    fonttitle=\bfseries,
    title=#1,
    arc=3mm,
    boxrule=1pt,
    drop fuzzy shadow={gray!50!white},
    left=5mm,
    right=5mm,
    top=3mm,
    bottom=3mm
}

\title{\textit{Thinking with Map}: \\
Reinforced Parallel Map-Augmented Agent for Geolocalization}

\newcommand{\project}{\raisebox{0pt}{\includegraphics[height=1.0em]{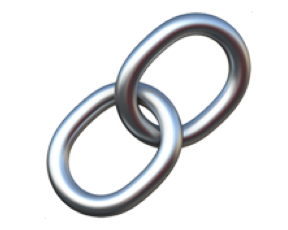}}\xspace}

\author{
 \textbf{Yuxiang Ji\textsuperscript{1,2}\setcounter{footnote}{0}\thanks{Work done during internship at AMAP, Alibaba Group.}} \quad
 \textbf{Yong Wang\textsuperscript{2}\setcounter{footnote}{1}\thanks{Project lead.}} \quad
 \textbf{Ziyu Ma\textsuperscript{2}} \quad
 \textbf{Yiming Hu\textsuperscript{2}} \quad 
 \textbf{Hailang Huang\textsuperscript{2}} \quad
\\
 \textbf{Xuecai Hu\textsuperscript{2}} \quad
 \textbf{Guanhua Chen\textsuperscript{3}} \quad
 \textbf{Liaoni Wu\textsuperscript{1}} \quad
 \textbf{Xiangxiang Chu\textsuperscript{2}}
\\
 \textsuperscript{1}Xiamen University\quad
 \textsuperscript{2}AMAP, Alibaba Group\quad
 \textsuperscript{3}Southern University of Science and Technology
\\
\\[-5pt]
\begin{tabular}{@{}ll@{}}
\vspace{.5em}\project\url{https://amap-ml.github.io/Thinking-with-Map}
\end{tabular}
\\
\\
}

\makeatletter
\AtBeginDocument{
\let\@oldmaketitle\@maketitle
\renewcommand{\@maketitle}{\@oldmaketitle
  \vspace{-13pt}
  \includegraphics[width=\linewidth]{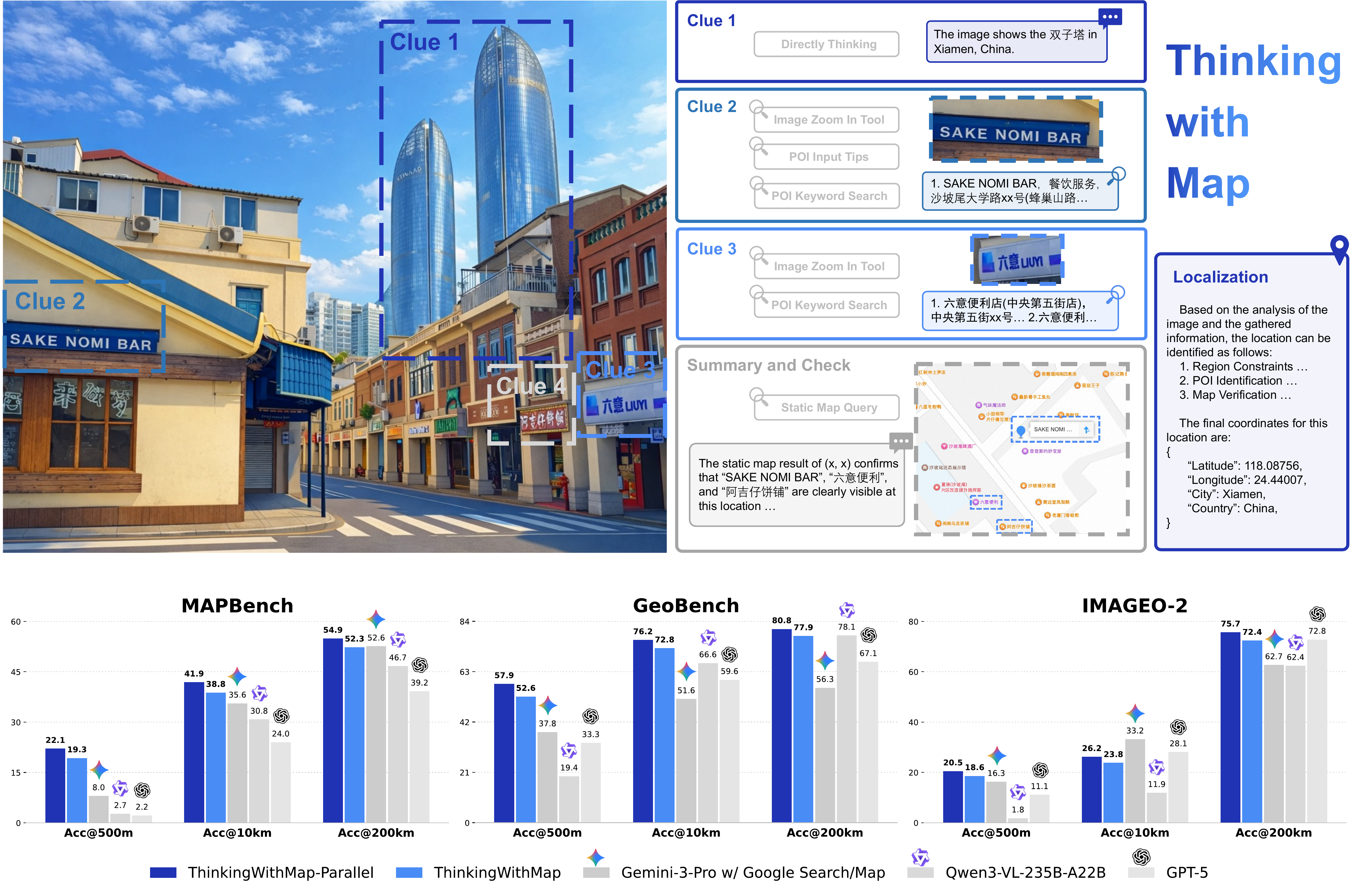}
  \vspace{-18pt}
  \captionof{figure}{
    \textbf{(Up)} The illustration of a complete \textit{Thinking with Map} process.
    \textbf{(Bottom)} Comparison with up-to-date open- and closed-source models on three geolocalization benchmarks.
    Our method is built upon the model \textit{Qwen3-VL-30B-A3B}.
    POI represents Point of Interest.
  }
  \label{fig:teaser}
  \vspace{17pt}
 }
}
\makeatother

\begin{document}
\maketitle

\begin{abstract}
The image geolocalization task aims to predict the location where an image was taken anywhere on Earth using visual clues.
Existing large vision-language model (LVLM) approaches leverage world knowledge, chain-of-thought reasoning, and agentic capabilities, but overlook a common strategy used by humans --- using maps.
In this work, we first equip the model \textit{Thinking with Map} ability and formulate it as an agent-in-the-map loop.
We develop a two-stage optimization scheme for it, including agentic reinforcement learning (RL) followed by parallel test-time scaling (TTS).
The RL strengthens the agentic capability of model to improve sampling efficiency, and the parallel TTS enables the model to explore multiple candidate paths before making the final prediction, which is crucial for geolocalization.
To evaluate our method on up-to-date and in-the-wild images, we further present MAPBench, a comprehensive geolocalization training and evaluation benchmark composed entirely of real-world images.
Experimental results show that our method outperforms existing open- and closed-source models on most metrics, specifically improving Acc@500m from 8.0\% to 22.1\% compared to \textit{Gemini-3-Pro} with Google Search/Map grounded mode.
\end{abstract}
\section{Introduction}

\begin{figure*}[t]
    \centering
    \includegraphics[width=\linewidth]{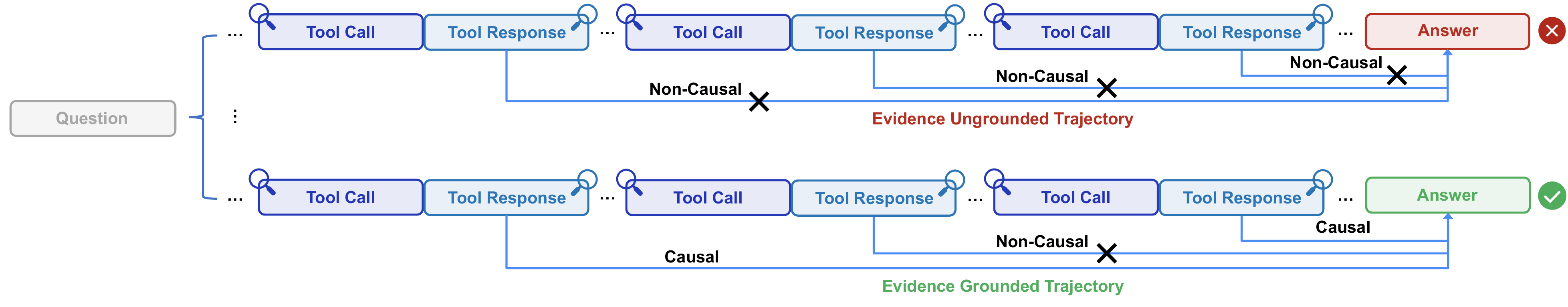}
    \caption{
        The \textit{Thinking with Map} trajectories from parallel sampling.
        The abundant map-API results make the trajectories easily verified based on their causal relationships.
    }
    \label{fig:parallel motivation}
\end{figure*}

Image geolocalization is a challenging task to determine the latitude and longitude of an image as accurately as possible.
Conventional vision research typically attributes this problem to a classification~\citep{seo_cplanet_2018,weyand_planet_2016,ferrari_geolocation_2018,clark_where_2023} or retrieval~\citep{ji2025game4loc,ji2025mmgeo,haas_pigeon_2024,yang_cross-view_2021,jia_g3_2024} task, achieving localization by predicting a region-level cell or retrieving the most similar image from a geo-tagged database.
Although these methods are well established in applications such as indoor localization~\citep{taira2018inloc,sarlin2019coarse} and landmark recognition~\citep{arandjelovic2016netvlad,noh2017large,weyand2020google,5206749}, they treat the entire image as a coupled feature for discrimination and fail to disentangle independent clues.
This less interpretable paradigm is inherently constrained by the training data and is difficult to generalize to images in the wild.

In the era of large vision-language models (LVLM), geolocalization can be viewed as a natural testbed for vision, understanding and reasoning.
Beyond single-image discriminative paradigm, it requires LVLMs to inspect visual clues (e.g., climate, architecture, and cultural context) in detail, and reason over the complex intersection of evidence to make the final prediction.
This process is closer to how human beings behave when inferring image locations.
Recent studies follow frontier models~\citep{deepseek-ai_deepseek-r1_2025,openai_o3mini_systemcard_2025,deepmind_gemini3pro_modelcard_2025,bai_qwen3-vl_2025,seed1.8_modelcard_2025,wang2025position} to further enhance such behavior by using chain-of-thought (CoT) reasoning~\citep{li2024georeasoner,li_recognition_2025,jia_geoarena_2025} and incorporating external tools within the reasoning chain~\citep{lai_mini-o3_2025,su2025thinking,qian_where_2025,wang_geovista_2025}.
However, despite their increased reasoning capability, these methods still depend on the model internal reasoning ability over knowledge.


In contrast, human beings rarely rely on internal reasoning alone for geolocalization.
When identifying visual clues, humans typically propose multiple location hypotheses and then verify them in turn using map tools.
By querying points of interest (POIs), examining road topology, and checking spatial consistency, maps provide an essential mechanism for validating visual clues against real-world geography.
Surprisingly, despite being the most fundamental tool for geolocalization, maps are almost absent from existing LVLM-based methods.
To bridge this gap, we equip the LVLM with map tools for the first time, enabling the model to \textit{Think with Map}.
Specifically, we expose map interfaces such as keyword search, POI details lookup, and static map query as callable tools, allowing the model to retrieve information and verify visual clues in the structured map environment during reasoning.
As illustrated in Figure~\ref{fig:teaser}, the process of \textit{Thinking with Map} is a multi-turn agentic behavior.
The model invokes tools based on multiple visual clues, then cross-validates the gathered evidence to produce the final prediction.
We further formulate this localization process as an agent-in-the-map loop, in which the agent iteratively proposes and verifies location hypotheses.

Similar to human beings, when the model encounters an ambiguous image, it needs to go through an iterative process of repeated hypothesis generation and verification.
However, simply increasing the reasoning budget to let the model explore sequentially not only leads to context explosion, but has also been found to yield marginal gains~\citep{wen_parathinker_2025,zheng_parallel-r1_2025}.
Inspired by the success of Google Gemini in parallel thinking~\citep{gemini_deep_think}, we also enable the model to explore multiple hypotheses in a parallel paradigm.
Unlike conventional reasoning tasks, \textit{Thinking with Map} inherently leaves a large number of map-API results in the reasoning trace.
These factual outputs make the reasoning trajectory largely self-verifying.
As Figures~\ref{fig:parallel motivation} and \ref{fig:parallel comparison}, we find that the LVLM can easily identify the better trajectory among multiple parallel \textit{Thinking with Map} trajectories by causal relationships.
Based on this observation, we introduce a simple parallel sampling with verifier framework for test-time scaling (TTS) in \textit{Thinking with Map}.
To further improve the model's pass@K performance and enable more effective parallel sampling, we conduct agentic Reinforcement Learning (RL) training for \textit{Thinking with Map}.

To evaluate our method, we propose MAPBench, which consists of up-to-date and broadly covered Chinese urban street-view and POI images.
We categorize the data into two difficulty levels for further analysis of the model's performance: \textit{easy} cases are those that the model can localize at a glance, while \textit{hard} cases contain less distinctive clues and are unlikely to be encountered during pre-training.
We also conduct rigorous evaluations on recently released benchmarks, including IMAGEO-Bench~\citep{li_pixels_2025} and GeoBench~\citep{wang_geovista_2025}.
The results show that our method consistently outperforms all open-source models by a large margin and even surpasses \textit{Gemini-3-Pro} (with Google Search/Map grounded mode) on most metrics.
Our contributions are summarized as follows:
\begin{itemize}[leftmargin=*,itemsep=2pt,topsep=0pt,parsep=0pt]
    \item We propose a map-augmented agent for the world-wide image geolocalization, equipped with the model \textit{Thinking with Map} ability.
    \item Building on the \textit{Thinking with Map} capability, we propose a parallel-and-verifier TTS method and further enhance it with agentic RL.
    \item We evaluate our method on the proposed MAPBench and other geolocalization benchmarks. The results show that our method outperforms all open- and closed-source models on most metrics.
\end{itemize}

\vspace{-0.1cm}
\section{Related Work}
\vspace{-0.1cm}

\textbf{Worldwide Geolocalization.}
Predicting the geographic location of a given image over the world is quite a challenging task~\citep{haas_pigeon_2024,qian_where_2025}.
Over the past decades, computer vision research primarily treats this task as a retrieval or classification problem.
The former relies on an enormous geo-tagged reference database as the retrieval gallery and introduces several large-scale benchmarks~\cite{hays2008im2gps,berton2022rethinking,berton2025megaloc}.
The latter partitions the Earth into structured ``geocells'' and predicts geographic coordinates either directly or hierarchically~\citep{ferrari_geolocation_2018,clark_where_2023,haas_pigeon_2024}.
Recent LVLM-based methods leverage the visual understanding and reasoning capabilities of frontier models to directly infer a location from an image, without any database or map partitioning~\citep{jia_geoarena_2025,li_recognition_2025,wang_llmgeo_2024,li_pixels_2025,huang_ai_2025}.
Although explicit reasoning reduces the black-box nature of the model, it cannot prevent hallucinations and biases of LVLMs.

\noindent\textbf{LVLM Powered Agent.}
As foundation models advance, researchers begin to focus on agentic capabilities and apply LVLM-powered agents to tasks that require interaction with open environments~\citep{team_kimi_2025,li_deepagent_2025,gur2023real,yao_react_2023}.
Recent works employ an end-to-end agentic RL~\citep{feng_group--group_2025,wang_ragen_2025,ji_tree_2025,dong_agentic_2025,chu_gpg_2025,yuan2025video,li2025adacurl,xiong2025hs} to improve tool use and long-horizon decision-making abilities of the base model in specific task environments, demonstrating a broad vision.
GeoVista~\citep{wang_geovista_2025} applies this paradigm to geolocalization by optimizing models to use vision and search tools for localization.
Some studies~\citep{qian_where_2025} also argue that general search tools offer very limited benefits for localization.
Beyond RL, some works also try to improve agent performance via test-time scaling methods such as parallel sampling~\citep{wen_parathinker_2025}, sequential revision~\citep{zhu_scaling_2025-1}, and multi-agent exploration~\citep{qiao_webresearcher_2025}.

\begin{figure*}[t]
    \centering
    \includegraphics[width=\linewidth]{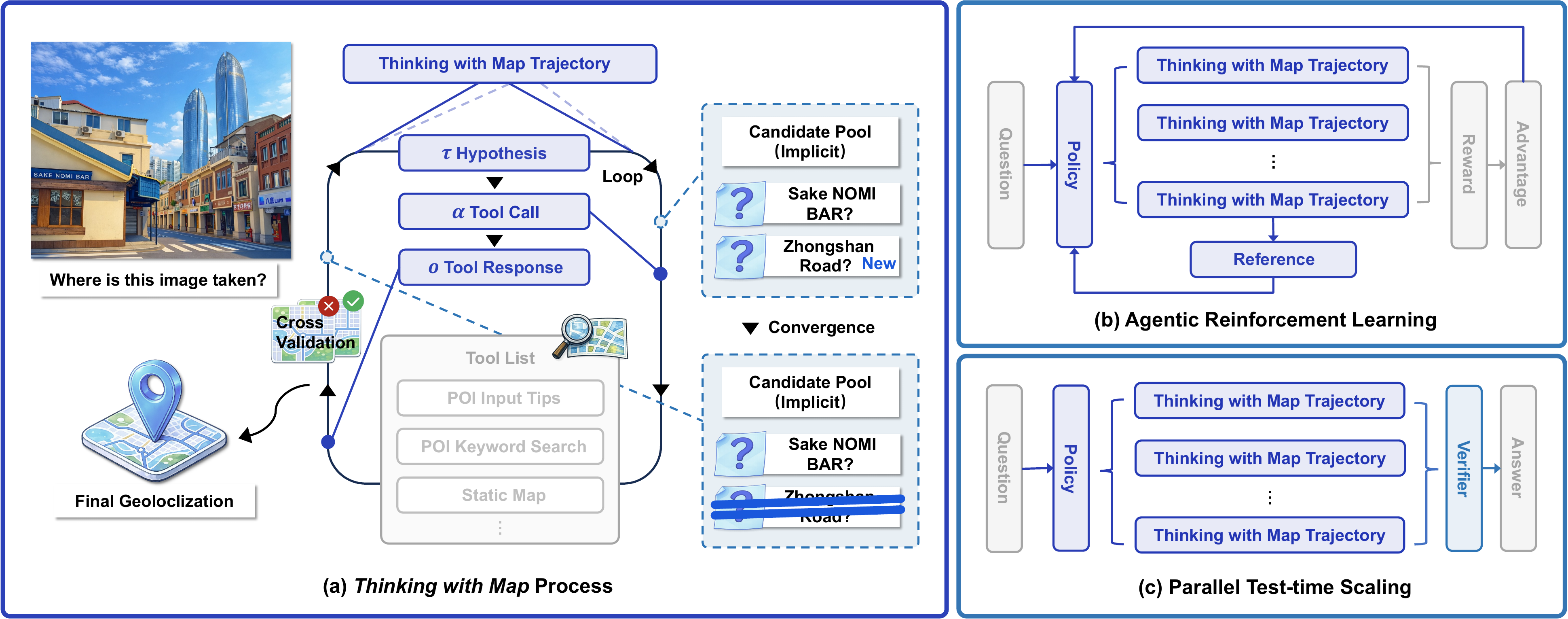}
    \caption{
        \textbf{(a)} The process of \textit{Thinking with Map}, consists of an agent-in-the-map loop.
        During the loop, the agent implicitly maintains a candidate pool of hypotheses.
        \textbf{(b)} The agentic reinforcement learning for \textit{Thinking with Map}.
        \textbf{(c)} The parallel test-time scaling with verifier pipeline for \textit{Thinking with Map}.
    }
    \label{fig:pipeline}
\end{figure*}

\begin{table}[t]
    \centering
    \resizebox{\linewidth}{!}{
    \begin{tabular}{l|c|c}
    \toprule
    \textbf{Tool Name} & \textbf{Parameter} & \textbf{Output} \\
    \midrule
    \texttt{image\_zoom\_tool} & Zoom in bounding box & Zoomed region image \\
    \texttt{poi\_input\_tips} & Query text & Search Suggestions \\
    \texttt{poi\_keyword\_search} & POI keyword & POI list \\
    \texttt{poi\_detail\_query} & POI id & POI details \\
    \texttt{static\_map\_query} & Location center & Static map image \\
    \texttt{satellite\_map\_query} & Location center & Satellite map image \\
    \bottomrule
    \end{tabular}
    }
    \caption{
        The involved tools for \textit{Thinking with Map}.
    }
    \vspace{-0.3cm}
    \label{tab:tool list}
\end{table}

\vspace{-0.1cm}
\section{Method}
\vspace{-0.1cm}
In this section, we present \textit{Thinking with Map}, a map-augmented agent for improved LVLM-based geolocalization.
The overview of our method can be viewed in Figure~\ref{fig:pipeline}.
We first present the definition and implementation of \textit{Thinking with Map} (\S~\ref{sec:thinking with map}).
Then we use agentic RL to improve sampling efficiency by optimizing performance from pass@N to pass@K (\S~\ref{sec:rl}).
Finally, we apply parallel TTS to explore multiple candidate hypotheses during geolocalization, to gain performance from pass@K to pass@1 (\S~\ref{sec:parallel tts}).

\vspace{-0.1cm}
\subsection{Thinking with Map}
\vspace{-0.1cm}
\label{sec:thinking with map}
Unlike direct discrimination or internal knowledge reasoning, we reformulate geolocalization as a \textit{Thinking with Map} process.
As Figure~\ref{fig:pipeline} (a), it follows an agent-in-the-map iterative loop of proposing location hypotheses, map retrieval, cross-validation and decision convergence.
Formally, we model \textit{Thinking with Map} as an iterative interaction process between a policy model $\pi_\theta$ and a structured map environment $P_\text{env}$.
Given a geolocalization query $q_\text{image,text}$, at each iteration $t$ the policy model can either propose a hypothesis $\tau_t$ (optional) explicitly/implicitly or verify existing hypotheses $\tau_{<t}$ through tool-call actions $\alpha_t$ to retrieve candidates within the map environment $P_\text{env}$.
Then the map tool responses are treated as an observation $o_t$, and together with all previous interaction history, form an evidence chain $s_t$ for cross-validation over the structured information:
\begin{equation}
    s_t = \{(\tau_0, \alpha_0, o_0), ...,(\tau_t, \alpha_t, o_t)\},
\end{equation}
\begin{equation}
\resizebox{\linewidth}{!}{$
    p_\theta(\tau,\alpha,o|s_0) = \prod_{t=0}^{T-1}\biggl[ \pi_{\theta}(\tau_t | s_t) \pi_{\theta}(\alpha_t | s_t, \tau_t) P_\text{env}(o_{t+1}|\alpha_t) \biggr].
$}
\end{equation}
Let there be an implicit candidate pool $\mathcal{C}_t$ in this iterative process.
Then the evidence chain $s_t$ composed by propositions and map observation at each step $t$ can be regarded as a maintenance update to the candidate pool:
\begin{equation}
    \mathcal{C}_{t+1} \triangleq \text{Update}(\mathcal{C}_{t}, s_t) \subseteq \mathcal{L},
\end{equation}
where $\mathcal{L}$ is the overall location set.
The policy model keeps maintaining this pool until it becomes sufficiently confident or the interaction budget is exhausted, and then selects the final answer from the candidate pool.


Here we provide a suite of map tools that human beings commonly use when looking for a location in Table~\ref{tab:tool list}.
Among these tools, POI search serves as the primary information source from the map engine, helping the model obtain location details for specific places.
Static and satellite maps then enable the model to verify and cross-check the surrounding scene and places around a candidate location.
Due to the region-specific availability of map services, we employ two types of map API providers\footnote{AMAP: \url{https://lbs.amap.com/}}\footnote{Google Map: \url{https://developers.google.com/maps}} to enable global geolocalization.
In addition, we provide an \texttt{image\_zoom\_tool}, which helps the model progressively inspect visual clues in large-scene images.

\begin{table*}[t]
    \centering
    \resizebox{\linewidth}{!}{
    \begin{tabular}{lccccccc}
    \toprule
    \textbf{Benchmark} & \textbf{Im2GPS3K} & \textbf{YFC100M} & \textbf{OSV-5M} & \textbf{IMAGEO-Bench} & \textbf{GeoBench} & \textbf{MAPBench} \\
    \midrule
    Reference & \citet{vo2017revisiting} & \citet{thomee2016yfcc100m} & \citet{astruc2024openstreetview} & \citet{li_pixels_2025} & \citet{wang_geovista_2025} & - \\
    Number & 3,000 & 100M & 5M & 6,152 / 2,929 / 220 & 512 / 512 / 108 & 2,500 / 2,500 \\
    Image Source & Flickr & Flickr & Mapillary & Mapillary/KartaView/Google Map & Web/Mapillary/Planetary Computer & AMAP \\ 
    Up-to-date & \ding{55} & \ding{55} & \ding{55} & \ding{55} & \ding{55} & \ding{51} \\ 
    Difficulty Tiering & \ding{55} & \ding{55} & \ding{55} & \ding{55} & \ding{55} & \ding{51} \\
    \bottomrule
    \end{tabular}
    }
    \caption{
        The comparison of MAPBench and existing geolocalization benchmarks.
    }
    \label{tab:dataset comp}
\end{table*}

\vspace{-0.1cm}
\subsection{RL for Map-augmented Agent}
\vspace{-0.1cm}
\label{sec:rl}
To enhance the model \textit{Thinking with Map} capability, we adopt a widely explored RL paradigm to improve agentic performance from pass@N to pass@K.
Instead of some recent \textit{Qwen2.5-VL}-based works~\citep{wang_geovista_2025,lai_mini-o3_2025,zheng2025deepeyes} that adopt a two-stage SFT-then-RL training pipeline, we find that the \textit{Qwen3-VL} model already shows basic tool-use ability after equipping it with map tools via the unified tool interface.
Therefore, we directly apply agentic RL from this base model.

As shown in Figure~\ref{fig:pipeline} (b), we adopt the  Group Relative Policy Optimization (GRPO)~\citep{shao_deepseekmath_2024} as the agentic RL algorithm.
Specifically, for each geolocalization query $q$, the LVLM-based agent generates a group of agent trajectories $\{\mathcal{H}_i = (\tau_0,\alpha_0,o_0,...,\tau_T,\alpha_T)\}_{i=1}^{G}$ based on the previous policy $\pi_{\theta_\text{old}}$.
The policy $\pi_\theta$ is then optimized by maximizing the advantages:
\begin{equation}
\label{eq:grpo object}
\begin{aligned}
    & J_\text{GRPO}(\theta) = \mathbb{E}_{q\sim \mathcal{D}, \mathcal{H} \stackrel{\textit{Agent}}{\sim} \pi_\text{old}(\cdot|q)} \Biggl[ \frac{1}{G} \sum_{i=1}^G \frac{1}{|\mathcal{H}^i|} \sum_{t=1}^{|\mathcal{H}^i|} \\ 
    & \quad \hat{r}_{i,t}(\theta) \hat{A}(\mathcal{H}^i) - \beta \mathbb{D}_\text{KL}\bigl(\pi_\theta(\mathcal{H}|q) \,\|\, \pi_\text{ref}(\mathcal{H}|q)\bigr)\Biggr],
\end{aligned}
\end{equation}
where $\hat{r}_{i,t}(\theta)$ is the importance sampling ratio, and clipping is applied in practice to stabilize RL training.
We prompt the model to output answers in a fixed JSON format for each query, enabling structured parsing for the verifiable reward function.
For geolocalization tasks evaluated by continuous distance, we simply use a piecewise discrete scheme that assigns different rewards to different distance ranges:
\begin{equation}
    r =
    \begin{cases}
    1, & dis \in [0, 500m)\\
    0.8, & dis \in [500m, 2km)\\
    0.6, & dis \in [2km, 10km)\\
    0.4, & dis \in [10km, 25km)\\
    0.2, & dis \in [25km, 200km)\\
    0.1, & dis \in [200km, 750km)\\
    0, & dis \in [750km, +\infty)\\
    \end{cases}
\end{equation}


This hierarchical reward reflects different localization granularity, e.g., $500m$ for fine-level and $25km$ for city-level.
In our experiments, this simple design works well with group-based RL and provides a discriminative learning signal.

\subsection{Parallel Test-time Scaling}
\label{sec:parallel tts}
After RL training, the reinforced model can perform image localization reasoning while interacting with map tools.
However, as with how human beings guess locations, images with limited clues often require a sequence of hypotheses and verification steps.
Due to the limited memory and reflection capabilities~\citep{li_deepagent_2025,liu_budget-aware_2025}, such long-horizon sequential reasoning is a challenging task for LVLM-based agents.

Fortunately, we find that \textit{Thinking with Map} trajectories naturally contain many self-verifiable factual information from map APIs, as shown in Figure~\ref{fig:parallel motivation}.
Therefore, we adopt a parallel-sampling pipeline with a verifier, where the model explores multiple paths through lightweight independent samples and a verifier aggregates the results.
Formally, given a geolocalization query $q$ and reinforced model $\pi_\theta$, we first sample a set of $N$ \textit{Thinking with Map} trajectories in parallel as:
\begin{equation}
\resizebox{\linewidth}{!}{$
    \biggl\{\mathcal{H} | \mathcal{H}_i = \prod_{t=0}^{T-1}\bigl[ \pi_{\theta}(\tau_t | s_t) \pi_{\theta}(\alpha_t | s_t, \tau_t) P_\text{env}(o_{t+1}|\alpha_t) \bigr]\biggr\}_{i=1}^{N}.
$}
\end{equation}
Then we feed the set of \textit{Thinking with Map} trajectories, together with the original image and a simple instruction $I$ into a LVLM-based verifier $\pi_\text{verifier}$, which summarizes the evidence and selects the most plausible prediction as:
\begin{equation}
\text{Answer} = \pi_\text{verifier} (q, \{\mathcal{H}\}_{i=1}^N, I).
\end{equation}

\begin{figure}[t]
    \centering
    \includegraphics[width=0.9\linewidth]{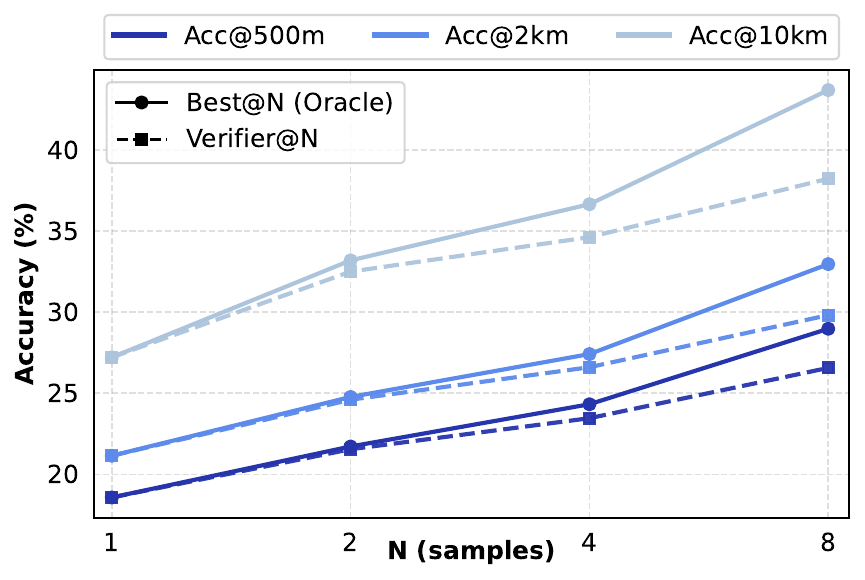}
    \caption{
        The comparison on parallel sampling.
    }
    \label{fig:parallel comparison}
\end{figure}

As Figure~\ref{fig:parallel comparison}, when we use \textit{Qwen3-VL-30B-A3B} to perform parallel sampling with different numbers, verifier@N closely matches oracle best@N.
In particular, when $N=2$ or $4$, the performance loss introduced by the verifier is almost negligible.
With this parallel test-time scaling, we enable the model to explore multiple \textit{Thinking with Map} hypotheses and aggregate self-verifiable trajectories to produce the final answer.
This approach transfers performance gains from pass@K to pass@1.

\begin{table*}[!t]
    \centering
    \resizebox{\textwidth}{!}{
    \begin{tabular}{lcccccccccccc}
        \toprule
        \multirow{2}{*}[-1em]{\textbf{Method}} & \multicolumn{6}{c}{\textbf{MAPBench-test-easy} ($Acc@Dis, \%$)} & \multicolumn{6}{c}{\textbf{MAPBench-test-hard} ($Acc@Dis, \%$)} \\
        \cmidrule(lr){2-7} \cmidrule(lr){8-13}
        & \makecell{Fine\\500m} & \makecell{Local\\2km} & \makecell{District\\10km} & \makecell{City\\25km} & \makecell{Region\\200km} & \makecell{Country\\750km} & \makecell{Fine\\500m} & \makecell{Local\\2km} & \makecell{District\\10km} & \makecell{City\\25km} & \makecell{Region\\200km} & \makecell{Country\\750km} \\
        \midrule
        \rowcolor{darkblue!10}
        \multicolumn{13}{c}{\textbf{\textit{Closed Source Model}}} \\
        \midrule
        GPT-o3 & 7.68 & 35.23 & 86.64 & 88.98 & 89.82 & 92.32 & 0.05 & 0.74 & 4.53 & 9.10 & 20.73 & 44.13 \\
        GPT-5 & 9.02 & 34.39 & \textbf{87.48} & \textbf{90.32} & \textbf{92.99} & \textbf{95.49} & 0.05 & 0.79 & 4.10 & 8.94 & 22.30 & 47.45 \\
        Gemini-3-Pro (w/. Google Search/Map) & 20.86 & 48.28 & 74.31 & 80.69 & 86.90 & 93.79 & 4.02 & 11.73 & 23.45 & 29.64 & 41.86 & 67.48 \\
        \midrule
        \rowcolor{darkblue!10}
        \multicolumn{13}{c}{\textbf{\textit{Open Source Model}}} \\
        \midrule
        Qwen3-VL-235B-A22B & 9.35 & 34.06 & 86.14 & 88.48 & 90.82 & 93.66 & 0.63 & 3.42 & 13.41 & 19.31 & 32.88 & 57.18 \\
        GLOBE-7B & 0.17 & 6.53 & 42.21 & 58.29 & 73.70 & 82.91 & 0.05 & 0.85 & 6.34 & 11.35 & 27.68 & 52.29 \\
        GeoVista-7B (w/. Google Search) & 0.33 & 4.17 & 28.21 & 39.39 & 47.74 & 51.08 & 0.00 & 0.53 & 4.16 & 6.52 & 10.94 & 18.99 \\
        Qwen3-VL-30B-A3B & 4.01 & 21.87 & 68.61 & 71.95 & 75.63 & 83.31 & 0.21 & 1.89 & 10.36 & 14.20 & 28.56 & 52.76 \\
        \; $+$ Thinking with Map & 33.10 & 40.28 & 53.68 & 56.89 & 59.94 & 64.73 & 10.83 & 12.05 & 16.08 & 19.06 & 25.58 & 38.28 \\
        \; $+$ Reinforcement Learning & 41.51 & 50.88 & 76.88 & 79.35 & 83.07 & 89.67 & 12.33 & 14.67 & 26.89 & 31.62 & 42.58 & 67.17 \\
        \; $+$ Parallel$\times 2$ \& Verifier & 43.65 & 54.38 & 79.93 & 82.27 & 85.12 & 90.64 & 13.70 & 16.45 & 28.98 & 33.79 & 44.32 & \textbf{68.85} \\
        \; $+$ Parallel$\times 4$ \& Verifier & \textbf{44.98} & \textbf{55.02} & 80.27 & 82.27 & 85.79 & 91.30 & \textbf{14.86} & \textbf{17.40} & \textbf{29.88} & \textbf{34.37} & \textbf{45.21} & \textbf{68.85} \\
        \bottomrule
    \end{tabular}
    }
    \caption{
        Comparison of \textit{Thinking with Map} with open- and closed-source models on MAPBench. 
        Results are reported as accuracy at multiple granularities ($Acc@Dis$).
        The \textbf{bold} indicates the best.
    }
    \label{tab:MAPBench test comparison}
\end{table*}

\section{Dataset}
As Table~\ref{tab:dataset comp}, most existing geolocalization benchmarks use images collected earlier from Google Street View~\citep{vo2017revisiting,wang2024llmgeo,wang_geovista_2025}, Mapillary~\citep{astruc2024openstreetview}, and Flickr~\citep{thomee2016yfcc100m}.
In our early experiments, we identified several major issues with these datasets:
\begin{itemize}[leftmargin=*,itemsep=2pt,topsep=0pt,parsep=0pt]
    \item \textbf{Timeliness.} Most of existing datasets are not up to date, and POIs shown in the images may no longer exist. As a result, they fail to assess an LVLM-based geolocalization method that leverage current, real-world knowledge. Moreover, obsolete POIs can contradict information from map APIs or the web, which can mislead the agent and impact localization performance.
    \item \textbf{Difficulty tiering.} Because LVLMs are pretrained on massive amount of world knowledge and images, many landmark-style images can be easily recognized and even memorized coordinates. Such images mainly measure memorization, but fail to evaluate the reasoning ability and capability to acquire and use external knowledge.
    \item \textbf{Global coverage.} Although existing datasets appear geographically diverse, their image sources bias them toward Europe and North America, with no coverage of China.
\end{itemize}

Based on these issues, we propose MAPBench, an up-to-date geolocalization benchmark with broad coverage across China.
The MAPBench consists of 5,000 nearby street-view images centered on POIs, with no POI repeated across samples.
We randomly split the dataset into 2,500 training samples and 2,500 test samples.
Furthermore, we categorize test samples based on the zero-shot predictions of three base models \textit{GPT-5}, \textit{GPT-o3}, and \textit{Qwen3-VL-235B-A22B}.
The sample is labeled as easy if at least two models predict locations within $10km$ of the ground truth, and labeled as hard otherwise.
The easy split evaluates the memorization and world knowledge of base model, while the hard split specifically assesses agentic capabilities.
As a result, 599 test samples are labeled as easy, while the remaining 1,901 test samples are labeled as hard.


\section{Experiment}

\begin{table*}[!t]
    \centering
    \resizebox{\textwidth}{!}{
    \begin{tabular}{lcccccccccccc}
        \toprule
        \multirow{2}{*}[-1em]{\textbf{Method}} & \multicolumn{6}{c}{\textbf{GeoBench} ($Acc@Dis, \%$)} & \multicolumn{6}{c}{\textbf{IMAGEO-2-test} ($Acc@Dis, \%$)} \\
        \cmidrule(lr){2-7} \cmidrule(lr){8-13}
        & \makecell{Fine\\500m} & \makecell{Local\\2km} & \makecell{District\\10km} & \makecell{City\\25km} & \makecell{Region\\200km} & \makecell{Country\\750km} & \makecell{Fine\\500m} & \makecell{Local\\2km} & \makecell{District\\10km} & \makecell{City\\25km} & \makecell{Region\\200km} & \makecell{Country\\750km} \\
        \midrule
        \rowcolor{darkblue!10}
        \multicolumn{13}{c}{\textbf{\textit{Closed Source Model}}} \\
        \midrule
        GPT-o3 & 33.08 & 50.75 & 61.99 & 64.45 & 67.67 & 73.45 & 9.66 & 18.76 & 27.41 & 30.85 & 47.06 & 67.04 \\
        GPT-5 & 33.30 & 46.90 & 59.64 & 63.17 & 67.13 & 75.05 & 11.14 & 19.91 & 28.12 & 32.62 & \textbf{50.62} & 72.78 \\
        Gemini-3-Pro (w/. Google Search/Map) & 37.79 & 47.22 & 51.61 & 53.64 & 56.32 & 59.10 & 16.33 & \textbf{27.33} & \textbf{33.22} & \textbf{37.00} & 48.78 & 62.67 \\
        \midrule
        \rowcolor{darkblue!10}
        \multicolumn{13}{c}{\textbf{\textit{Open Source Model}}} \\
        \midrule
        Qwen3-VL-235B-A22B & 19.38 & 46.68 & 66.60 & 71.52 & 78.05 & \textbf{91.54} & 1.78 & 5.66 & 11.88 & 15.76 & 34.07 & 62.38 \\
        GLOBE-7B & 11.21 & 43.69 & 69.15 & 71.72 & 78.50 & 88.78 & 0.33 & 1.33 & 4.77 & 7.77 & 31.74 & 65.37 \\
        GeoVista-7B (w/. Google Search) & 6.85 & 26.55 & 45.50 & 51.17 & 54.81 & 58.35 & 0.22 & 1.11 & 3.77 & 5.66 & 12.54 & 20.08 \\
        Qwen3-VL-30B-A3B & 12.21 & 40.47 & 66.60 & 71.52 & 76.02 & 90.90 & 1.11 & 3.22 & 8.77 & 12.99 & 34.52 & 65.82 \\
        \; $+$ Thinking with Map & 49.82 & 59.05 & 66.64 & 68.28 & 71.72 & 81.36 & 17.75 & 19.33 & 21.55 & 23.72 & 31.93 & 47.36 \\
        \; $+$ Reinforcement Learning & 52.57 & 64.01 & 72.83 & 74.53 & 77.92 & 86.62 & 18.64 & 20.50 & 23.77 & 27.19 & 42.59 & 72.41 \\
        \; $+$ Parallel$\times 2$ \& Verifier & 55.61 & 67.06 & 75.23 & 76.17 & 79.44 & 87.38 & 19.64 & 21.86 & 25.53 & 29.08 & 45.06 & 74.14 \\
        \; $+$ Parallel$\times 4$ \& Verifier & \textbf{57.94} & \textbf{69.16} & \textbf{76.17} & \textbf{77.57} & \textbf{80.84} & 89.02 & \textbf{20.53} & 22.64 & 26.19 & 30.19 & 46.06 & \textbf{75.69} \\
        \bottomrule
    \end{tabular}
    }
    \caption{
        Comparison of \textit{Thinking with Map} with open- and closed-source models on GeoBench and IMAGEO. 
        Results are reported as accuracy at multiple granularities ($Acc@Dis$).
        The \textbf{bold} indicates the best.
    }
    \label{tab:GeoBench IMAGEO test comparison}
\end{table*}

\begin{table}[!t]
    \centering
    \resizebox{\linewidth}{!}{
    \begin{tabular}{lcccccc}
        \toprule
        \multirow{2}{*}[-1em]{\textbf{RL Method}} & \multicolumn{6}{c}{\textbf{MAPBench-test-all} ($Acc@Dis, \%$)} \\
        \cmidrule(lr){2-7}
        & \makecell{Fine\\500m} & \makecell{Local\\2km} & \makecell{District\\10km} & \makecell{City\\25km} & \makecell{Region\\200km} & \makecell{Country\\750km} \\
        \midrule
        Qwen3-VL-30B-A3B & 1.12 & 6.67 & 24.29 & 28.01 & 39.82 & 60.07 \\
        \; $+$ \texttt{image\_zoom\_tool} & 1.48 & 6.81 & 23.27 & 26.53 & 35.36 & 53.60 \\
        \; $+$ \texttt{web\_search\_tool} & 1.77 & 9.55 & 26.05 & 29.34 & 36.73 & 49.73 \\
        \; $+$ \texttt{map\_tool} & 16.16 & 18.80 & 25.07 & 28.11 & 33.80 & 44.61 \\
        \bottomrule
    \end{tabular}
    }
    \caption{
        The ablation study on tool types.
    }
    \vspace{-0.3cm}
    \label{tab:tool ablation}
\end{table}

\subsection{Experimental Setup}
\textbf{Models.}
We compare the proposed \textit{Thinking with Map} against multiple series of state-of-the-art closed-source models, including \textit{GPT-o3} and \textit{GPT-5} from OpenAI, and \textit{Gemini-3-Pro} from Google.
We also compare against a large-scale open-source model \textit{Qwen3-VL-235B-A22B} from Alibaba, as well as two open-source geolocalization methods GLOBE~\citep{li_recognition_2025} and GeoVista~\citep{wang_geovista_2025}.
Our method is built upon \textit{Qwen3-VL-30B-A3B-Instruct}.

\noindent\textbf{Datasets.}
To evaluate our method for worldwide geolocalization capability, in addition to the proposed MAPBench, we also include two recently released benchmarks IMAGEO-Bench~\citep{li_pixels_2025} and GeoBench~\citep{wang_geovista_2025}.
In particular, we use an IMAGEO-2 subset as it exhibits greater difficulty in our experiments.
For RL training, we use the MAPBench training set and 2,000 examples from IMAGEO-2, achieving globally covered samples.
More details are in Appendix~\ref{sec:dataset detail}.

\noindent\textbf{Evaluation.}
To analyze the model localization accuracy at different granularities, we report $acc@dis$ at six levels (500m@Fine, 2km@Local, 10km@District, 25km@City, 200km@Region and 750km@Country), with distance thresholds matching the reward settings.
Specifically, a prediction is considered correct if its distance to the ground truth is below the corresponding threshold.

\noindent\textbf{Settings.}
For closed-source models, we query them directly via APIs.
Some of them have built-in tool-use capabilities, such as image manipulation tools of \textit{GPT-o3} and Google Search / Google Maps grounded mode of \textit{Gemini-3-Pro}.
For the two open-source geolocalization methods, we follow the original papers to set the corresponding inference hyperparameters, and equip GeoVista-7B with \texttt{image\_zoom\_tool} and \texttt{web\_search\_tool} via a unified tool interface.
If not specified, we use \textit{Qwen3-VL-235B-A22B} as the verifier for the results of parallel sampling.
More details are in Appendix~\ref{sec:implementation details}.

\begin{figure*}[t]
    \centering
    \includegraphics[width=\linewidth]{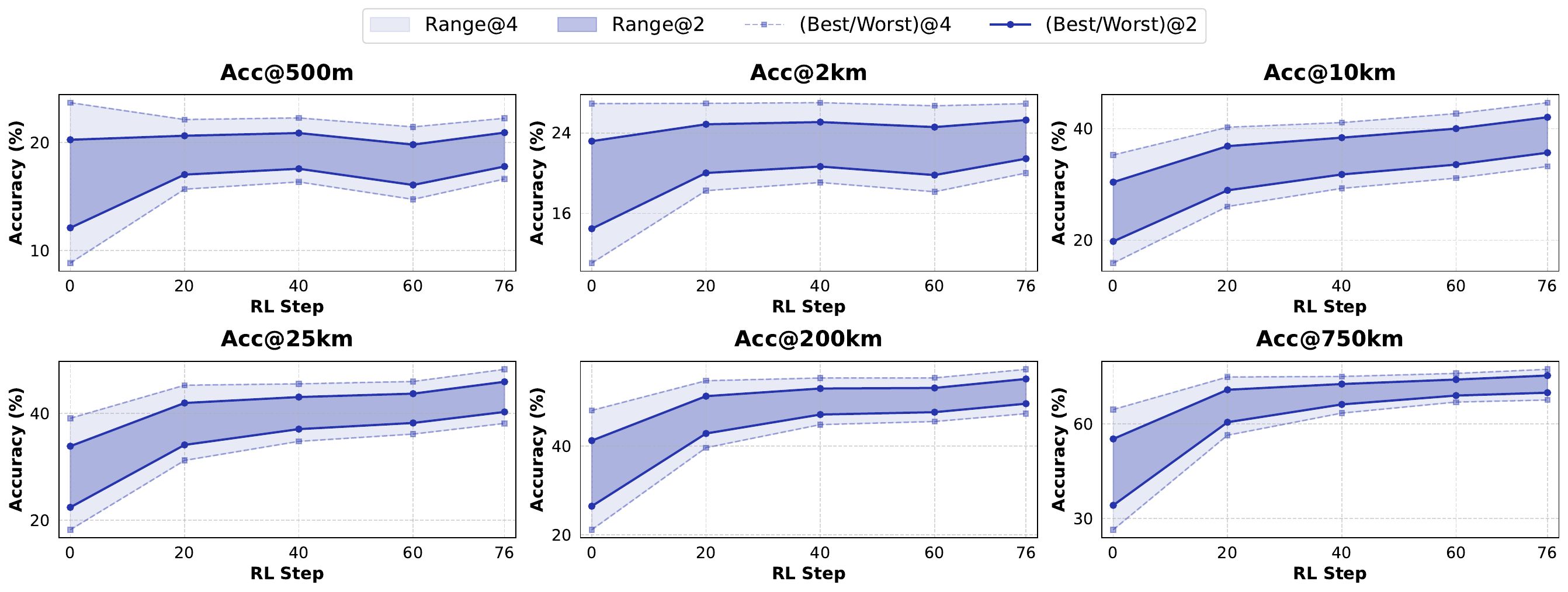}
    \vspace{-0.7cm}
    \caption{
        The evolution of pass@K accuracy across RL training steps on MAPBench.
    }
    \label{fig:passk rl evolution}
\end{figure*}

\begin{table*}[!t]
    \centering
    \resizebox{\linewidth}{!}{
    \begin{tabular}{lcccccccccccc}
        \toprule
        \multirow{2}{*}[-1em]{\textbf{Verifier Model}} & \multicolumn{6}{c}{\textbf{MAPBench-test-easy} ($Acc@Dis, \%$)} & \multicolumn{6}{c}{\textbf{MAPBench-test-hard} ($Acc@Dis, \%$)}\\
        \cmidrule(lr){2-7} \cmidrule(lr){8-13} 
        & \makecell{Fine\\500m} & \makecell{Local\\2km} & \makecell{District\\10km} & \makecell{City\\25km} & \makecell{Region\\200km} & \makecell{Country\\750km} & \makecell{Fine\\500m} & \makecell{Local\\2km} & \makecell{District\\10km} & \makecell{City\\25km} & \makecell{Region\\200km} & \makecell{Country\\750km} \\
        \midrule
        \rowcolor{darkblue!10}
        \multicolumn{13}{c}{\textbf{\textit{Verifier@2}}} \\
        \midrule
        Qwen3-VL-30B-A3B & 43.48 & 53.18 & 77.93 & 80.27 & 83.95 & 89.97 & 13.64 & 16.34 & 28.34 & 32.95 & 42.99 & 67.42 \\
        Qwen3-VL-235B-A22B & 43.65 & 54.35 & 79.93 & 82.27 & 85.12 & 90.64 & 13.70 & 16.45 & 28.98 & 33.79 & 44.32 & 68.86 \\
        GPT-5 & 43.81 & 54.01 & 79.93 & 82.27 & 85.79 & 91.47 & 13.86 & 16.61 & 28.45 & 33.21 & 43.89 & 68.06 \\
        \midrule
        \rowcolor{darkblue!10}
        \multicolumn{13}{c}{\textbf{\textit{Verifier@4}}} \\
        \midrule
        Qwen3-VL-30B-A3B & 44.15 & 53.85 & 79.26 & 81.10 & 85.12 & 90.13 & 14.65 & 17.03 & 28.98 & 33.74 & 44.32 & 68.48 \\
        Qwen3-VL-235B-A22B & 44.98 & 55.02 & 80.27 & 82.27 & 85.79 & 91.30 & 14.86 & 17.40 & 29.88 & 34.37 & 45.21 & 68.85 \\
        GPT-5 & 45.82 & 54.85 & 80.94 & 83.11 & 86.96 & 92.31 & 14.86 & 17.19 & 29.88 & 34.58 & 44.79 & 68.96 \\
        \bottomrule
    \end{tabular}
    }
    \caption{
        The ablation study on verifier models.
        Verifier@N means verifier with N parallel samples.
    }
    \label{tab:verifier ablation}
\end{table*}

\subsection{Main Results}
\label{sec:main results}
As shown in Tables~\ref{tab:MAPBench test comparison} and \ref{tab:GeoBench IMAGEO test comparison}, our proposed \textit{Thinking with Map} method achieves the best performance comparing with all open- and closed-source models on most metrics across four test sets.
In particular, for fine localization Acc@500m, our method outperforms the best closed-source model \textit{Gemini-3-Pro} on MAPBench-test-hard by a large margin, from 4.02\% to 14.86\%.
The substantial gains on GeoBench and IMAGEO-2-test also show improvingAcc@500m from 37.79\% to 57.94\% and 16.33\% to 20.53\%, respectively.
Due to the base model used in existing open-source geolocalization methods are relatively small (7B), their performance also cannot match closed-source models.
On the other hand, our task directly predicts latitude and longitude, which differs from the models original training targets and can hurt performance.

In our experiments, we find that the capability of base model can determine coarse-grained localization performance (e.g., Acc@25km and Acc@200km), while the search and map tools can greatly enhance fine-grained localization performance (e.g., Acc@500m).
For example, on MAPBench-test-hard, all base models achieve nearly 0\% accuracy for fine-localization, while only \textit{Gemini-3-Pro} with Google Search/Map grounded mode and our method reach 4.02\% and 14.86\% respectively.
However, directly integrating map tools can also lead to negative effects.
Noisy information from the map tools (e.g., wrong search results) may introduce substantial bias in coarse localization, which is reflected by the performance drop in ``$+$ Thinking with Map'' row.
This performance drop is addressed after reinforcement learning training.
Notably, our \textit{Thinking with Map} method already outperforms the other approaches even before incorporating parallel TTS.

When incorporating parallel TTS, our method achieves further performance gains, and the improvement is positively correlated with the number of parallel samples.
This gain trend is consistent with that of the base model with parallel TTS in Figure~\ref{fig:parallel comparison}.

\subsection{Quantitative Analysis}
\textbf{Different Tools.}
Here we explore how different types of tools affect the geolocalization task.
We use \textit{Qwen3-VL-30B-A3B-Instruct} as the base model and integrate three types of tools separately.
The results in Table~\ref{tab:tool ablation} align with our earlier discussion in \S~\ref{sec:main results}.
All three tool types improve fine-grained localizaiton ($<2km$), but hurt coarse-grained localization ($>200km$).
Among them, \texttt{image\_zoom\_tool} and \texttt{web\_search\_tool} bring very marginal improvements, whereas \texttt{map\_tool} yields a clear gain from 1.12\% to 16.16\% on Acc@500m.

\noindent\textbf{Evolution of Pass@K across RL.}
Many recent studies~\citep{yue2025does} explore the impact of RL-based post-training on LVLMs.
Here we evaluate the effect of RL on the geolocalization task by examining the evolution of pass@K accuracy throughout RL training, as shown in Figure~\ref{fig:passk rl evolution}.
As RL training progresses, the prediction accuracy at all granularities shows lower variance, as Range@2/4 becoming smaller.
This trend is consistent with the view that RL helps optimize performance from pass@K toward pass@1.
Notably, accuracy at larger distance thresholds (i.e., $Dis>10km$) shows a clear upward trend under best@N.
This suggests that RL also helps the model achieve stronger pass@K from pass@N ($K<N$).
However, Best@500m shows little to no improvement, and can even limit exploration.

\noindent\textbf{Different Verifier Models.}
To further validate the role of the verifier and investigate what makes a better verifier in parallel TTS, we experiment with different verifier models in Table~\ref{tab:verifier ablation}.
The results show that when the parallel size $N=2$, the choice of model has only a minor impact, and a 30B model is already sufficient to serve as a strong verifier.
As the parallel size increases, the verifying task becomes harder, and the impact of model capacity becomes correspondingly more important.

\vspace{-0.2cm}
\section{Conclusion}
\vspace{-0.1cm}
In this work, we propose a map-augmented agent for image geolocalization, to enable model \textit{Thinking with Map}.
We model this process as an agent-in-the-map loop of proposing hypotheses, map retrieval, cross-validation, and decision convergence.
Based on this, we propose a two-stage optimizaiton approach that combines agentic RL and parallel test-time scaling to gain pass@N capability within a single query.
Experimental results show that our method outperforms all open- and closed-source models on most metrics.

\section*{Limitation}
In this work, we equip the agent with map tools, enabling the LVLM agent to do geolocalization by iteratively interacting within a structured map environment.
Although the model can perform evidence-grounded reasoning with map tools, we find that its map-use ability still falls far short of human performance. 
For example, we do not observe the model inferring orientation from relative spatial relationships, which is a common strategy humans use when estimating locations.
For agentic RL, our training data remain very limited, which constrains the model to learn in open environments.
One promising avenue for future work is to investigate what emergent capabilities arise when scaling up this RL paradigm.
Finally, we consider parallel TTS a pragmatic interim solution that compensates for the current limitations of a single agent.
How to build a single agent with stronger long-horizon problem-solving capabilities remains an open problem.
\section{Acknowledgment}
We acknowledge the helpful discussion with Kaibin Tian, the author of SeekWorld~\citep{seekworld2025}, and our intern colleagues, Shidong Yang and Zengbin Wang for their assistance.

\bibliography{custom}

\appendix
\section*{Appendix}

\section{Datasets}
\label{sec:dataset detail}
Here we provide more details of our proposed MAPBench.
We uniformly and randomly sample 5,000 valid POIs across 20 cities in China, and for each POI we randomly select either a street-view or storefront photo, forming a final set of 5,000 images.
This simple construction process ensures that the samples are both up-to-date and broadly coverage.


Considering the worldwide coverage and timeliness of the image sources, in addition to our proposed MAPBench, we also use two recently released datasets for global images:
\begin{itemize}[leftmargin=*,itemsep=2pt,topsep=0pt,parsep=0pt]
    \item \textbf{IMAGEO-2} is a subset of IMAGEO-Bench~\citep{li_pixels_2025}, and constructed from crowdsourced images from Google Map POIs. The original data are released by \citet{yan2023personalized}, then compiled and filtered to final 2,929 images. We use 2,027 randomly sampled instances for training (as IMAGEO-2-train) and the remaining 902 instances for testing (as IMAGEO-2-test).
    \item \textbf{GeoBench}~\citep{wang_geovista_2025} is a recently released datasets composed of three types images, including 512 normal photos, 512 panoramas and 108 satellite images. The normal photos are sourced from Internet, the panoramas are collected via the Mapilary API, and the satellite images come from Sentinel-2 Level-2A imagery accessed through Microsoft Planetary Computer. We use all the data for testing.
\end{itemize}

\begin{table}[t]
    \centering
    \begin{tabular}{lc}
        \toprule
        Config & Setting \\
        \midrule
        \rowcolor{darkblue!10}
        \multicolumn{2}{c}{\textit{RL Training}} \\
        \midrule
        optimizer & AdamW \\
        learning rate & 1e-6 \\
        KL coefficient & 0.001 \\
        training epoch & 2 \\
        training batch size & 64 \\
        PPO mini batch size & 16 \\
        max response length & 4096 \\
        max tool response length & 1024 \\
        max turns & 8 \\
        group size & 16 \\
        \midrule
        \rowcolor{darkblue!10}
        \multicolumn{2}{c}{\textit{Parallel Testing}} \\
        \midrule
        top K & 60 \\
        top P & 0.95 \\
        temperature & 1.0 \\
        \bottomrule
    \end{tabular}
    \caption{
        Hyperparameters for \textit{Thinking with Map} RL training and parallel testing.
    }
    \label{tab:hyperparameters}
\end{table}

\begin{table*}[!t]
    \centering
    \resizebox{\linewidth}{!}{
    \begin{tabular}{lcccccccccccc}
        \toprule
        \multirow{2}{*}[-1em]{\textbf{Verifier Model}} & \multicolumn{6}{c}{\textbf{GeoBench} ($Acc@Dis, \%$)} & \multicolumn{6}{c}{\textbf{IMAGEO-2-test} ($Acc@Dis, \%$)} \\
        \cmidrule(lr){2-7} \cmidrule(lr){8-13}
        & \makecell{Fine\\500m} & \makecell{Local\\2km} & \makecell{District\\10km} & \makecell{City\\25km} & \makecell{Region\\200km} & \makecell{Country\\750km} & \makecell{Fine\\500m} & \makecell{Local\\2km} & \makecell{District\\10km} & \makecell{City\\25km} & \makecell{Region\\200km} & \makecell{Country\\750km} \\
        \midrule
        \rowcolor{darkblue!10}
        \multicolumn{13}{c}{\textbf{\textit{Verifier@2}}} \\
        \midrule
        Qwen3-VL-30B-A3B & 56.78 & 66.82 & 75.47 & 76.40 & 79.44 & 87.38 & 19.76 & 21.75 & 25.19 & 28.41 & 45.62 & 74.25 \\
        Qwen3-VL-235B-A22B & 55.61 & 67.06 & 75.23 & 76.17 & 79.44 & 87.38 & 19.64 & 21.86 & 25.53 & 29.08 & 45.06 & 74.14 \\
        GPT-5 & 60.51 & 72.90 & 80.37 & 81.31 & 84.11 & 90.65 & 21.64 & 24.20 & 28.52 & 31.63 & 49.06 & 75.69 \\
        Best@2 & 57.48 & 69.86 & 77.34 & 78.27 & 80.84 & 88.79 & 19.76 & 22.09 & 26.42 & 30.52 & 48.72 & 78.36 \\
        \midrule
        \rowcolor{darkblue!10}
        \multicolumn{13}{c}{\textbf{\textit{Verifier@4}}} \\
        \midrule
        Qwen3-VL-30B-A3B & 57.71 & 69.86 & 76.64 & 77.80 & 81.07 & 89.02 & 20.31 & 22.09 & 25.97 & 29.41 & 45.84 & 74.14 \\
        Qwen3-VL-235B-A22B & 57.94 & 69.16 & 76.17 & 77.57 & 80.84 & 89.02 & 20.53 & 22.64 & 26.19 & 30.19 & 46.06 & 75.69 \\
        GPT-5 & 63.32 & 75.00 & 82.01 & 83.64 & 86.45 & 92.76 & 22.09 & 24.64 & 29.19 & 33.07 & 49.39 & 77.36 \\
        Best@4 & 61.92 & 73.13 & 78.50 & 79.44 & 82.48 & 89.95 & 22.31 & 24.42 & 28.52 & 33.74 & 53.05 & 82.24 \\
        \bottomrule
    \end{tabular}
    }
    \caption{
        The ablation study of verifier models on GeoBench and IMAGEO.
        Verifier@N means verifier with N parallel samples.
    }
    \label{tab:more verifier ablation}
\end{table*}

\begin{figure}[!t]
    \centering
    \includegraphics[width=\linewidth]{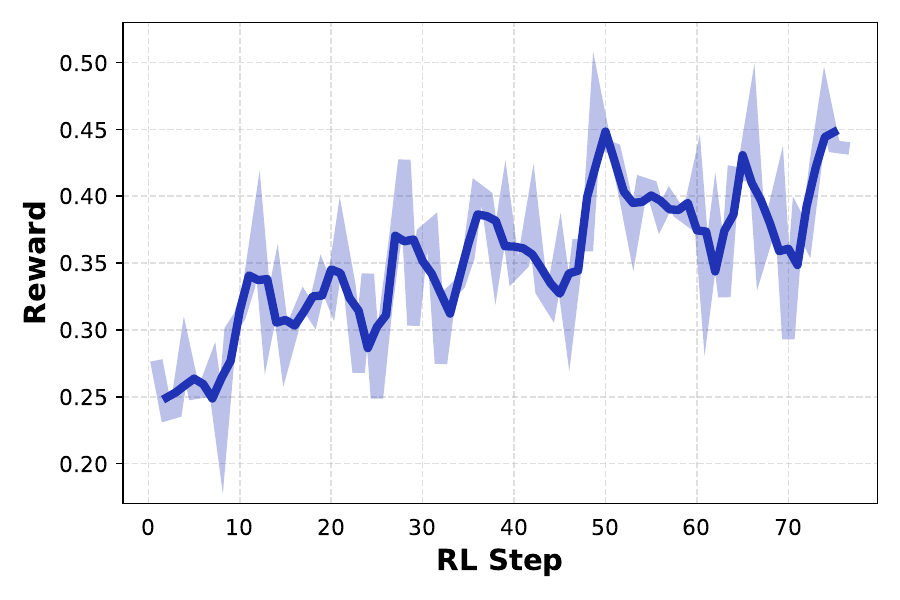}
    \caption{
        Reward dynamics across RL training.
    }
    \label{fig:rl reward}
\end{figure}

\section{Experiment Details}

\subsection{Implementation Details}
\label{sec:implementation details}

Our agentic RL training is implemented on VeRL codebase.
The specific hyperparameter settings for RL training and parallel testing are shown in Table~\ref{tab:hyperparameters}.
The RL training and other experiments are conducted on 32 NVIDIA H20 GPUs.

\begin{table}[!t]
    \centering
    \resizebox{\linewidth}{!}{
    \begin{tabular}{lcccccc}
        \toprule
        \multirow{2}{*}[-1em]{\textbf{RL Algorithm}} & \multicolumn{6}{c}{\textbf{AMAP-test-all} ($Acc@Dis, \%$)} \\
        \cmidrule(lr){2-7}
        & \makecell{Fine\\500m} & \makecell{Local\\2km} & \makecell{District\\10km} & \makecell{City\\25km} & \makecell{Region\\200km} & \makecell{Country\\750km} \\
        \midrule
        GRPO & \textbf{19.33} & \textbf{23.36} & \textbf{38.89} & \textbf{43.07} & \textbf{52.29} & \textbf{72.57} \\
        Pass@K-GRPO & 16.28 & 19.35 & 27.52 & 31.91 & 40.46 & 60.48 \\
        PKPO & 16.97 & 19.35 & 26.43 & 30.15 & 36.68 & 50.41 \\
        \bottomrule
    \end{tabular}
    }
    \caption{
        The ablation study of RL algorithm.
    }
    \label{tab:ablation of RL algorithm}
\end{table}

Here we provide the prompt template for \textit{Thinking with Map} and other base models as follows.
They all pose a straightforward geolocalization task and require the final answer to be returned in a fixed JSON format.
The only difference is that the former additionally provides guidance on tool use.
The verifier prompt consists of the original geolocalization query together with multiple parallel \textit{Thinking with Map} trajectories.
The prediction format matches the requirements for the single-agent and base-model setting, that the answer must be in the same fixed JSON format.

\subsection{Training Dynamics of RL}
\label{sec:training dynamics of rl}
To better understand the benefits of agentic RL, we show the reward dynamics over RL training steps in Figure~\ref{fig:rl reward}.
From the reward curve, we find that the training reward increases from 0.25 in the early stage to 0.45 by the end, showing an overall upward trend.
This further demonstrates the positive effect of RL on localization accuracy.
In the second epoch (i.e., the latter half of training), the reward gradually oscillates and approaches to stable, which suggests that more data may be needed.

\subsection{Ablation Study on RL Algorithm}
\label{sec:ablation on rl algorithm}
We also try other RL algorithms for \textit{Thinking with Map} agentic training, in particular Pass@K-GRPO~\citep{tang2025optimizing} and PKPO~\citep{walder2025pass}.
Results in Table~\ref{tab:ablation of RL algorithm} show that although these methods explicitly optimize for pass@K, they perform substantially worse than vanilla GRPO on our task.
Therefore, we still use GRPO-trained model for parallel TTS.

\subsection{More Ablation Studies on Verifier Models}
\label{sec:more ablation on verifier}
Here we provide more ablation studies of verifier models on GeoBench and IMAGEO-2-test.
As shown in Table~\ref{tab:more verifier ablation}, unlike the results on MAPBench, using a verifier based on a different base model (e.g., GPT-5) can even outperform the corresponding Best@N (Oracle).
This suggests that the verifier is not merely selecting among existing candidates.
In few cases, it also identifies more plausible answers along the \textit{Thinking with Map} trajectory.

\begin{templatebox}{Prompt Template for Thinking with Map}
<image>You are given an image, and your task is to use your exceptional skills to determine the precise coordinates of the location depicted. 

Carefully examine the image, taking note of any distinctive features, POIs, landmarks, vegetation, or other elements that could serve as clues. 

When extra information is needed to search for a location or confirm precise coordinates, you can use the given tools to get the information from search engine and maps. 

Once you have gathered sufficient evidence, provide your best inference for the coordinates in the following JSON format: 

\{"lat": latitude, "lon": longitude, "city": city, "country": country\}.

Use signed values for latitude and longitude to indicate N/S and E/W.
If you cannot narrow it down, then provide your best guess.
\end{templatebox}

\begin{templatebox}{Prompt Template for Base Model}
<image>You are given an image, and your task is to use your exceptional skills to determine the precise coordinates of the location depicted.

Carefully examine the image, taking note of any distinctive features, POIs, landmarks, vegetation, or other elements that could serve as clues.

After showing your thinking, provide your final answer in the JSON format:

\{"lat": latitude, "lon": longitude, "city": city, "country": country\}

Use signed values for latitude and longitude to indicate N/S and E/W.
If you cannot narrow it down, then provide your best guess.
\end{templatebox}

\begin{templatebox}{Prompt Template for Verifier}
You are a strict geo-localization solver.

You will be given an image, the original task, and multiple candidate answers from other agents.
Synthesize the best final location.

If candidates disagree, pick the most evidence-consistent and geographically plausible one. 

After thinking, provide your final answer in the JSON format: 

\{"lat": latitude, "lon": longitude, "city": city, "country": country\}.

Use signed values for latitude and longitude to indicate N/S and E/W.
\end{templatebox}


\end{document}